\documentclass[runningheads]{llncs}
\usepackage[T1]{fontenc}
\usepackage{booktabs}
\usepackage{amssymb}
\usepackage{graphicx}
\usepackage{xcolor}
\begin{document}
\title{ExtractGPT: Exploring the Potential of Large Language Models for Product Attribute Value Extraction}
%\author{Anonymous}
\author{Alexander Brinkmann\inst{1}\orcidID{0000-0002-9379-2048} \and
Roee Shraga\inst{2}\orcidID{0000-0001-8803-8481} \and
Christian Bizer\inst{1}\orcidID{0000-0003-2367-0237}}
\authorrunning{A. Brinkmann et al.}
%\authorrunning{Anonymous}
\titlerunning{ExtractGPT}

\institute{University of Mannheim, 68131 Mannheim, Germany \\
\and
Worcester Polytechnic Institute, Worcester Polytechnic Institute\\
\email{\{alexander.brinkmann,christian.bizer\}@uni-mannheim.de}\\
\email{rshraga@wpi.edu}}
\maketitle              % typeset the header of the contribution
\begin{abstract}
%Comment by Alex: The abstract must not exceed 150 words.
%After Chris' edits: 147 words
E-commerce platforms require structured product data in the form of attribute-value pairs to offer features such as faceted product search or attribute-based product comparison. However, vendors often provide unstructured product descriptions, necessitating the extraction of attribute-value pairs from these texts. BERT-based extraction methods require large amounts of task-specific training data and struggle with unseen attribute values. This paper explores using large language models (LLMs) as a more training-data efficient and robust alternative. We propose prompt templates for zero-shot and few-shot scenarios, comparing textual and JSON-based target schema representations. Our experiments show that GPT-4 achieves the highest average F1-score of 85\% using detailed attribute descriptions and demonstrations. Llama-3-70B performs nearly as well, offering a competitive open-source alternative. GPT-4 surpasses the best PLM baseline by 5\% in F1-score. Fine-tuning GPT-3.5 increases the performance to the level of GPT-4 but reduces the model's ability to generalize to unseen attribute values.

\end{abstract}

\section{Introduction}
\label{sec:introduction}

\begin{figure}[ht]
\centering
\includegraphics[width=.6\textwidth]{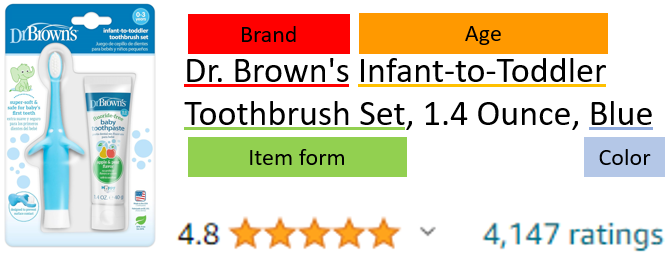}
\caption{An example product title with tagged attribute-value pairs. Vendors include product attribute values in the title to enhance visibility.}
\label{fig:task_teaser}
\end{figure}

Online shoppers often filter and compare products using criteria such as brand, color, or screen size to find products that fit their needs~\cite{ren_information_2018}.
However, many vendors only provide textual product descriptions~\cite {wang_learning_2020,zhu_multimodal_2020}. To enable a precise faceted search for products, attribute-value pairs need to be extracted from these textual product descriptions~\cite{yang_mave_2022,zheng_opentag_2018}. 
Figure \ref{fig:task_teaser} shows an example offer for a toothbrush set including attribute-value annotations. In the figure, attribute names are marked with colored boxes. Attribute values are underlined.

State-of-the-art techniques for attribute value extraction (AVE) often rely on pre-trained language models (PLMs)~\cite{chen_does_2023,wang_learning_2020,xu_scaling_2019,yang_mave_2022,zhu_multimodal_2020} such as BERT~\cite{devlin_bert_2019}.
However, these methods require large amounts of task-specific training data and struggle to generalize to unseen attribute values.
Large language models (LLMs) like GPT-3.5~\cite{ouyang_training_2022}, GPT-4~\cite{openai_gpt-4_2023} and Llama-3~\cite{dubey_llama_2024} have proven effective in mitigating these shortcomings for various natural language processing tasks~\cite{brown_language_2020,wei_emergent_2022}. 
This paper proposes and evaluates prompt templates for instructing LLMs to extract attribute values in zero-shot and few-shot scenarios. We test different approaches to represent the target attributes for the zero-shot scenario. In the few-shot scenario, we evaluate three methods for using training data: (i) providing example attribute values, (ii) selecting in-context demonstrations, and (iii) fine-tuning the LLM. We compare the performance of the best prompt template/LLM combination with state-of-the-art PLM-based methods fine-tuned with varying amounts of training data.
The contributions of the paper are as follows:

%\rs{Currently all "contributions" we declare are empirical. I would also add at least one conceptual contribution, e.g., "We offer a LLM prompting scheme for Product attribute/value pairs extraction using zero, few-shot and fine-tuning..."}

\begin{enumerate}
    \item We propose different prompt templates for instructing LLMs about the target attribute schema of the AVE task. The templates cover use cases with and without task-specific training data.  
    \item Our experiments show that LLMs require a small set of task-specific training data for picking example values and/or demonstrations to reach decent performance.
    \item Our comparison of the different approaches to use training data shows that providing demonstrations is more effective than providing example values. 
    \item We show that GPT-4 outperforms all other LLMs with a top average F1 score of 85\%. The best open-source LLM, Llama-3-70B, has an F1 score only 3\% lower than GPT-4's top score, making it a strong open-source alternative.
    \item Comparing LLMs and PLMs, we show that LLMs are more training data-efficient. Given the same amount of training data, GPT-4 achieves a 5\% higher average F1 score and is 19\% better with unseen values than the best PLM-based method AVEQA.    
    \item Our experiments show that a fine-tuned GPT-3.5 model has a similar performance as GPT-4 but loses some of its ability to generalize to unseen attribute values. 
    %Our experiments show that starting with 1.4k extracted attribute/value pairs, it is more cost-effective to fine-tune a GPT-3.5 model than to use GPT-4 with prompts that include example values and demonstrations.
\end{enumerate}
%\rs{I like the contributions but IMHO, reviewers may be overwhelmed with the number. My suggestion is: (a) combine contributions (1) and (2) are general, (3) and (4) present what we found/suggest about prompting, and (5) and (6)  (potentially also 7) which refer to the results. Obviously this is not a must consider combining one/two.}

The paper is structured as follows. First, we review related work. Next, we describe the experimental setup (Section~\ref{sec:experimental_setup}) before delving into prompt engineering. We introduce zero-shot prompt engineering (Section~\ref{sec:prompt_engineering}) as well as in-context learning and fine-tuning (Section~\ref{sec:training_data}) to evaluate the usage of task-specific training data. Lastly, we compare LLM- and PLM-based methods in Section~\ref{sec:baselines}. 
The code and data for replicating our experiments are available online\footnote{https://github.com/wbsg-uni-mannheim/ExtractGPT}.

\section{Related Work}

\vspace{.1cm}\noindent\textbf{Attribute Value Extraction.}
The goal of AVE is to extract specific attribute values from unstructured text, such as product titles and descriptions, based on a pre-defined target schema, which represents a set of target attributes.
Early works used domain-specific rules to extract attribute-value pairs~\cite{vandic_faceted_2012,zhang_framework_2009} from product descriptions.
Initial learning-based methods required extensive feature engineering and did not generalize to unknown attributes and values~\cite{ghani_text_2006,putthividhya_bootstrapped_2011,wong_scalable_2009}.
Recent approaches use BiLSTM-CRF architectures, with OpenTag~\cite{zheng_opentag_2018} and its extension SU-OpenTag~\cite{xu_scaling_2019} utilizing BERT~\cite{devlin_bert_2019} for encoding. AdaTag~\cite{yan_adatag_2021} employs BERT~\cite{devlin_bert_2019} and a mixture-of-experts module for AVE. 
Recently, many works approach AVE as a question-answering task, using different PLMs to encode the target attribute, product category, and product title~\cite{shinzato_simple_2022,wang_learning_2020,yang_mave_2022}.
The PLM-based models SU-OpenTag, AVEQA and MAVEQA serve as baselines for our experiments.
OA-Mine~\cite{zhang_oa-mine_2022} employs BERT~\cite{devlin_bert_2019} to mine for unknown attributes and values.
%Recent studies have utilized soft prompt tuning to fine-tune a small number of trainable parameters in a language model~\cite{blume_generative_2023,yang_mixpave_2023}.
Brinkmann et al. use LLMs to extract and normalize attribute values~\cite{brinkmann_using_2024}.
LLM-ensemble employs an ensemble of LLMs for AVE~\cite{fang_llm-ensemble_2024}.
Other works deal with AVE from multiple modalities like texts and images~\cite{wang_smartave_2022,zhu_multimodal_2020}.

\vspace{.1cm}\noindent\textbf{Information Extraction using LLMs.}
%LLMs often show better zero-shot performance compared to PLMs and are more robust to unseen examples~\cite{brown_language_2020} because they are pre-trained on large amounts of text, and have emergent effects due to their model size~\cite{wei_emergent_2022}.
LLMs have successfully been used for information extraction tasks in other application domains:
Wang et al.~\cite{wang_code4struct_2023} and Parekh et al.~\cite{parekh_geneva_2023} employed OpenAI's LLMs to extract structured data about events from unstructured text.
Goel et al.~\cite{goel_llms_2023} combine LLMs with human expertise to annotate patient-related information in medical texts.
Khorashadizadeh et al.~\cite{khorashadizadeh_exploring_2023} explore in-context learning for LLMs to generate knowledge graphs from text.

%Agrawal et al.~\cite{agrawal_large_2022} use InstructGPT with zero-shot and few-shot prompts to extract information from clinical notes.
%Shyr et al.~\cite{shyr_identifying_2023} evaluate ChatGPT to extract rare disease phenotypes from unstructured text. 
%LLMs have been used to re-rank information extracted by PLM-based models~\cite{ma_large_2023}.
%In the next section, we provide the experimental setup to discuss our proposed methods.
\section{Experimental Setup}
\label{sec:experimental_setup}

Our main research question is how to instruct an LLM to extract the values of the attributes mentioned in the target schema.
This section presents our setup to approach this question experimentally, including datasets, LLMs, and evaluation metrics.

\subsection{Datasets}

We use the benchmark datasets OA-Mine~\cite{zhang_oa-mine_2022} and AE-110K~\cite{xu_scaling_2019}, which have been used by related work~\cite{wang_learning_2020,yang_mixpave_2023,yang_mave_2022}. Both datasets consist of English product offers with annotated attribute-value pairs. The MAVE~\cite{yang_mave_2022} dataset is not considered, because its attribute-value pairs are annotated using an ensemble of fine-tuned AVEQA models~\cite{wang_learning_2020} and are not manually verified, hence during human-inspection we found numerous annotation errors~\cite{zou_implicitave_2024}. Other datasets used in the related work are not publicly available~\cite{yan_adatag_2021,zheng_opentag_2018}.

\sloppy
\vspace{.1cm}\noindent\textbf{OA-Mine.}
We use the human-annotated product offers of the OA-Mine dataset\footnote{https://github.com/xinyangz/OAMine/tree/main/data}~\cite{zhang_oa-mine_2022}. The subset includes 10 product categories, with up to 200 product offers per category. Each category has between 8 and 15 attributes, resulting in a total of 115 unique attributes. Attributes with the same name but different product categories are treated as distinct attributes. We do not apply any further pre-processing to the subset of OA-Mine.  

\vspace{.1cm}\noindent\textbf{AE-110K.}
The AE-110K dataset\footnote{https://raw.githubusercontent.com/lanmanok/ACL19\_Scaling\_Up\_Open\_Tagging/\linebreak master/publish\_data.txt} comprises triples of product titles, attributes and attribute values from the AliExpress Sports \& Entertainment category~\cite{xu_scaling_2019}. Product offers are derived by grouping the triples by title. The subset contains 10 categories, with up to 400 offers per category. For each category, 6 to 17 attributes are known, resulting in a total of 101 unique attributes.

\vspace{.1cm}\noindent\textbf{Training/Test Split.} The subsets OA-Mine and AE-110K are split into training and test sets in a 75:25 ratio, stratified by product category to ensure the presence of all attributes in both sets. To evaluate the impact of the amount of training data on LLM performance, we create small and large training sets for OA-Mine and AE-110K. The large training sets include all available training data. For the small training set, 20\% of the product offers per category are sampled from the available training data.
Table \ref{tab:dataset_statistics} presents statistics for OA-Mine and AE-110K.

\begin{table}[ht]
\centering
\caption{Statistics for OA-Mine and AE-110K. }
\label{tab:dataset_statistics}
\begin{tabular}{@{}l|r|r|r|r|r|r@{}}
\toprule  
                        & \multicolumn{3}{l|}{OA-Mine}                                                          & \multicolumn{3}{l}{AE-110K}                                                          \\ \midrule
                        & Small & Large &  & Small & Large &  \\
                        & Train & Train & Test & Train & Train & Test \\ \midrule
%Unique Cat.       & 10                          & 10                          & 10                       & 10                          & 10                          & 10                       \\
%Unique Attr.       & 115                         & 115                         & 115                      & 101                         & 101                         & 101                      \\
Attribute-Value Pairs   & 1,467                       & 7,360                       & 2,451                    & 859                         & 4,360                       & 1,482                    \\
Unique Attribute-Value Pairs & 1,120                       & 4,177                       & 1,749                    & 302                         & 978                         & 454                      \\
Product Offers          & 286                         & 1,452                       & 491                      & 311                         & 1,568                       & 524   \\ \bottomrule             
\end{tabular}
\end{table}
%\begin{example}[Example Extractions]
\vspace{.1cm}\noindent\textbf{Example Extractions.} 
    Table \ref{tab:example_extractions} shows example product offer titles, target attributes and attribute values from the datasets. The attribute values are \underline{underlined} in the titles. The lower part of the table shows the attribute values extracted by GPT4 using the prompt templates \texttt{list}, \texttt{json-val} and \texttt{json-val-dem} which will be introduced in Section~\ref{sec:prompt_engineering} and Section~\ref{sec:training_data}. 
    The extraction of attribute values is either correct (\textcolor{green}{\checkmark}) or incorrect (\textcolor{red}{$\times$}).
%\end{example}

\begin{table*}[t]
\caption{Example product offers, attribute-value pairs and extracted attribute values for the two datasets OA-Mine and AE-110k.}
\label{tab:example_extractions}
\begin{tabular}{l|l|l|l}
\toprule
\textbf{Dataset}        & OA-Mine                    & OA-Mine          & AE-110k                \\
\textbf{Category}        & Vitamin                    & Coffee           & Guitar                 \\
\textbf{Attribute}       & Net Content                & Flavor           & Body Material          \\ \midrule
\textbf{Title}           & NOW Supplements,           & Cafe Du Monde    & Factory customization  \\
                & Vitamin A (Fish Liver Oil) & \underline{Coffee Chicory}, & Acoustic Guitar Sitika \\
                & 25,000 IU, Essential       & 15-Ounce         & \underline{Solid Spruce} Vintage   \\
                & Nutrition, \underline{250} Softgels    & (Pack of 3)      & Sunburst high-quality  \\ \midrule
\textbf{Target Value}   & 250    & Coffee Chicory  & Solid Spruce            \\ \midrule
\textbf{list}            & n/a    \textcolor{red}{$\times$}                    & n/a \textcolor{red}{$\times$}            & Sitika Solid Spruce \textcolor{red}{$\times$}    \\
\textbf{json-val}      & 250 Softgels \textcolor{red}{$\times$}               & n/a \textcolor{red}{$\times$}             & Solid Spruce \textcolor{green}{\checkmark}         \\
\textbf{json-val-dem} & 250 \textcolor{green}{\checkmark}    & Coffee Chicory \textcolor{green}{\checkmark}  & Solid Spruce \textcolor{green}{\checkmark}  \\ \bottomrule
 \multicolumn{1}{r}{}& \multicolumn{1}{c}{(a)} & \multicolumn{1}{c}{(b)} & \multicolumn{1}{c}{(c)} \\
\end{tabular}
\end{table*}

\subsection{Large Language Models}
\sloppy
This paper evaluates the proprietary LLMs GPT-3.5 and GPT-4 as well as the open-source LLMs Llama-3-8B and Llama-3-70B~\cite{dubey_llama_2024}. Table \ref{tab:llms} lists the LLMs with the exact model name, number of parameters, and access via API or number of GPUs for running the LLM locally. 
We access GPT-3.5 and GPT-4 through the OpenAI API. The open-source models Llama-3-8B\footnote{https://huggingface.co/meta-llama/Meta-Llama-3-8B-Instruct} and Llama-3-70B\footnote{https://huggingface.co/meta-llama/Meta-Llama-3-70B-Instruct} are publicly available and were run on local GPUs.
The temperature parameter of all LLMs is set to 0 to reduce the randomness.
%The OpenAI API and the local execution of open-source models are facilitated using the langchain Python package\footnote{https://python.langchain.com/en/latest/index.html}. 
We use up to four NVIDIA RTX A6000 GPUs to run the open-source LLMs and fine-tune the PLM-based baselines.

\begin{table}[h]
\centering
\caption{List of evaluated LLMs.}
\label{tab:llms}
\begin{tabular}{@{}llll@{}}
\toprule
LLM     & Exact Name   & Model Size                 & API/GPUs                  \\ \midrule
GPT-3.5~\cite{ouyang_training_2022}   & gpt-3.5-turbo-0613 & 175B                       & API                     \\
GPT-4~\cite{openai_gpt-4_2023}     & gpt-4-0613         & unknown & API                     \\
%Beluga2   & StableBeluga2      & 70B                        & \multicolumn{1}{r}{4} \\
%Beluga-7B & StableBeluga-7B    & 7B                         & \multicolumn{1}{r}{1} \\
%Solar     & SOLAR-0-70b-16bit  & 70B                        & \multicolumn{1}{r}{3} \\ 
Llama-3-8B & Meta-Llama-3-8B-Instruct    & 8B                         & \multicolumn{1}{r}{1} \\
Llama-3-70B & Meta-Llama-3-70B-Instruct    & 70B                         & \multicolumn{1}{r}{4} \\\bottomrule
\end{tabular}
\end{table}

\subsection{Evaluation Metrics}

We report the F1-score, calculated by categorizing predictions into five groups as per previous works~\cite{wang_learning_2020,xu_scaling_2019,yan_adatag_2021,yang_mave_2022}. The five categories are NN (no predicted value, no ground truth value), NV (predicted value, no ground truth value), VN (no predicted value, ground truth value), VC (predicted value matches ground truth value), and VW (predicted value does not match ground truth value). The F1-score is derived from the precision ($P = VC / (NV + VC + VW)$), recall ($R = VC/ (VN + VC + VW)$), and the formula $F1 = 2PR/(P + R)$.

\section{Zero-Shot Prompt Engineering}
\label{sec:prompt_engineering}

This section discusses a zero-shot scenario, meaning that no training data is available. The main research question in this scenario is: how to describe the target schema to the LLMs, specifically, how to define which attributes should be extracted.
First, we introduce the structure of our prompt templates. Afterwards, we analyze the representation of the target schema along two dimensions: the level of detail of the attribute descriptions and the format in which the target schema is presented.

\begin{figure*}[ht]
\centering
\includegraphics[width=.9\textwidth]{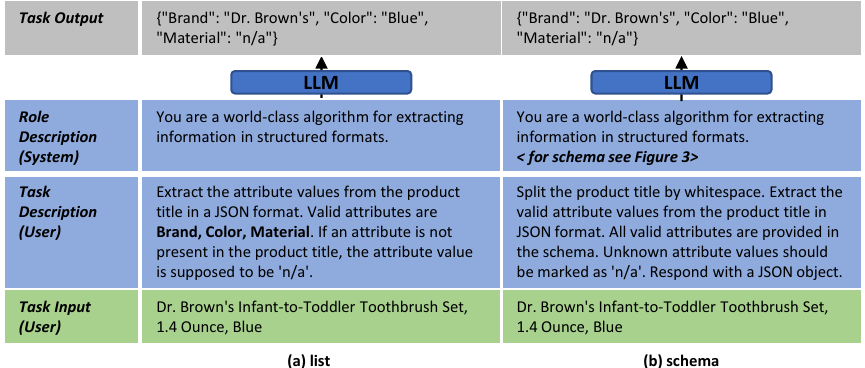}
\caption{Zero-shot prompt templates \texttt{list} and \texttt{schema}.}
\label{fig:zero_shot_prompts}
\end{figure*}

\subsection{Prompt Templates}
\label{sub_sec:prompt_structure}
Prompt design is key for having LLMs reach a high performance~\cite{zamfirescu-pereira_why_2023}. Thus, we create various prompt templates to systematically analyze different target attribute representations. All templates instruct the LLM to simultaneously extract attributes from product titles, leveraging synergies between them.
Each template consists of four chat messages: role description (\textcolor{blue}{blue}), task description (\textcolor{blue}{blue}), task input (\textcolor{green}{green}) and task output (\textcolor{gray}{grey}), as shown in Figure \ref{fig:zero_shot_prompts}. The role description outlines the LLM's behaviour. The task description provides AVE instructions, including pre-processing the product title, formatting the response as JSON, and marking absent attributes as n/a. The task input contains the product title. Role and task description are static whereas task input and task output change with each extraction. 
The chat prompt template combines role description, task description, and task input, with each message having a specific type: \texttt{system} for the role description, and \texttt{user} for the task description and input. The LLM's response, the task output, is supposed to be a JSON document adhering to the target schema, enabling the evaluation of the task output.

\subsection{Representation of the Target Schema}
\begin{figure}[ht]
\centering
\includegraphics[width=.9\textwidth]{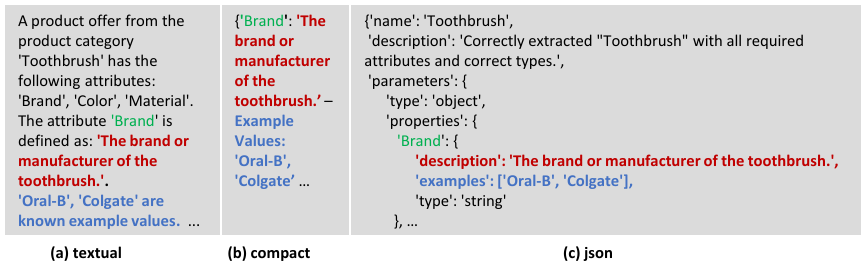}
\caption{Target schema representation formats (a) \texttt{textual}, (b) \texttt{compact} and (c) \texttt{json}.}
\label{fig:schema_types}
\end{figure}

This section investigates how different target schema representations affect the performance of LLMs on the attribute value extraction task. The representations are evaluated based on two dimensions: (i) level of detail and (ii) representation format.
Depending on the level of detail an attribute in the target schema is described by its name, a description, and example values. We evaluate four representation formats:
\begin{itemize}
    \item \texttt{list} enumerates the attribute names as illustrated in Figure \ref{fig:zero_shot_prompts} (a).
    \item \texttt{textual} articulates names (\textcolor{green}{green}), descriptions (\textcolor{red}{red}) and example values in plain text as depicted in Figure \ref{fig:schema_types} (a).
    \item \texttt{compact} densely combines names (\textcolor{green}{green}), descriptions (\textcolor{red}{red}) and example values (\textcolor{blue}{blue}) as shown in Figure \ref{fig:schema_types} (b).
    \item \texttt{json} represents names (\textcolor{green}{green}), descriptions (\textcolor{red}{red}) and example values (\textcolor{blue}{blue}) using the JSON schema vocabulary\footnote{https://json-schema.org/} as depicted in Figure \ref{fig:schema_types} (c).
\end{itemize}

The product category determines the set of attributes in the target schema.
On average, OA-Mine and AE-110k define 10 attributes per product category.
For demonstration purposes, Figure \ref{fig:zero_shot_prompts} and Figure \ref{fig:schema_types} display a subset of the attributes for the 'toothbrush' category.
The \texttt{schema} prompt template integrates the \texttt{textual}, \texttt{compact} and \texttt{json} schema representations into the role description as seen in Figure \ref{fig:zero_shot_prompts}. 
We generate attribute descriptions using GPT-3.5 since the original datasets do not provide descriptions.
In the zero-shot scenario, example values are absent as they need to be selected from training data. Section \ref{sub_sec:example_values} covers the use of example values. 

\begin{table}[h]
\centering
\caption{F1-scores for zero-shot prompt templates. The highest F1-score per dataset is marked in bold.}
\label{tab:prompt_engineering_zero_shot}
\begin{tabular}{@{}l|rrrr|rrrr@{}}
\toprule
& \multicolumn{4}{c|}{OA-Mine} & \multicolumn{4}{c}{AE-110K} \\ \midrule
Model & list & textual & compact & json & list & textual & compact & json \\ \midrule
 GPT-3.5 & 63.3 & 62.7 & 65.1 & 64.8 & 61.4 & 61.3 & \textbf{63.6} & 61.5 \\
 GPT-4 & \textbf{69.1} & 68.9 & 68.8 & 68.1 & 56.3 & 55.5 & 53.7 & 62.1 \\
 Llama-3-8B & 62.9 & 65.2 & 64.6 & 43.2 & 57.8 & 57.7 & 58.1 & 22.0 \\
 Llama-3-70B & 67.3 & 64.3 & 63.9 & 49.9 & 58.9 & 52.0 & 56.0 & 30.0 \\
\bottomrule
\end{tabular}
\end{table}

As shown in Table~\ref{tab:prompt_engineering_zero_shot}, overall, GPT-3.5, GPT-4 and the Llama-3 LLMs achieve zero-shot F1-scores above 60\%. None of the target schema representations has a clear advantage over the other representations across all LLMs. The \texttt{json} representation is difficult to interpret for all open-source, while for GPT-3.5 and GPT-4 this drop in performance is not observed. Once we add example values in Section \ref{sec:training_data}, GPT-3.5 and GPT-4 benefit most from the \texttt{json} representation.
%The open-source models perform worse than OpenAI's LLMs on all prompts and do not benefit from the target schema representations with great detail. The \texttt{list} prompt works best for open-source models. 
% \rs{consider using the `example' environment. It typically helps readers (reviewers) navigate the paper more easily.}
%\begin{example}
%    Looking at the extraction examples for the \texttt{list} prompt template in Table \ref{tab:example_extractions}, we see that GPT-4's background knowledge about brands is sufficient for correctly extracting the brand name `PENERAN' in examples (d). On the other hand, GPT-4 does not correctly predict the attribute values of the examples (a), (b) and (c) using the \texttt{list} prompt template.
%\end{example}

\section{Using Training Data}
\label{sec:training_data}
This section explores the scenario where training data is available. We investigate using the training data for (i) adding example values to the target schema representations, (ii) sampling demonstrations for in-context learning, and (iii) fine-tuning GPT-3.5.

\subsection{Example Values}
\label{sub_sec:example_values}

This section explores the effect of adding attribute values from the training set to the target schema representations introduced in Section 
\ref{sec:prompt_engineering}. 
The evaluation is conducted in two steps. First, we compare the representation formats \texttt{textual}, \texttt{compact}, and \texttt{json} with example values (\texttt{val}) to the best representation without example values (\texttt{compact}). Second, we assess how different amounts of sampled attribute values affect the performance of the best representation.
By default, up to 10 unique values are randomly sampled for each attribute from the training set. If less than 10 unique values are found, all available values are retrieved.

\vspace{.1cm}\noindent\textbf{Target Schema Representations.}
Table~\ref{tab:prompt_engineering_with_example_values} shows the impact of adding attribute values to the target schema representations. The average F1 performance of GPT and Llama-3 LLMs increases by 5\% to 16\% compared to the \texttt{compact} representation without attribute values. OpenAI's LLMs perform best with the \textit{json} representation, while Llama-3 LLMs excel with the \textit{compact} representation.

\begin{table}[h]
\centering
\caption{F1-scores for prompt templates with example values. The highest F1-score per dataset is marked in bold.}
\label{tab:prompt_engineering_with_example_values}
\begin{tabular}{@{}l|r|rrr|r|rrr@{}}
\toprule
 & \multicolumn{4}{c|}{OA-Mine} & \multicolumn{4}{c}{AE-110K} \\ \midrule
 & 0-val &\multicolumn{3}{c|}{10-val} & 0-val & \multicolumn{3}{c}{10-val} \\
Model & compact & compact & textual & json & compact & compact & textual & json \\ \midrule
GPT-3.5 & 65.1 & 60.7 & 61.4 & 69.8 & 63.6 & 55.0 & 40.5 & \textbf{74.4} \\
GPT-4 & 68.8 & 64.8 & 66.4 & \textbf{75.0} & 53.7 & 53.5 & 46.5 & 70.1 \\
Llama-3-8B & 64.6 & 67.5 & 70.9 & 25.5 & 58.1 & 69.9 & 70.6 & 20.3 \\
Llama-3-70B & 63.9 & 71.1 & 73.0 & 70.3 & 56.0 & 66.0 & 64.0 & 50.0 \\
\bottomrule
\end{tabular}
\end{table}

\vspace{.1cm}\noindent\textbf{Amount of example values.}
We now evaluate the impact of the number of unique example values sampled from the training set on the performance of GPT-3.5 and GPT-4. Samples of 3, 5, and 10 example values per attribute are taken.
For each configuration, we calculate the total number of sampled unique attribute values (Unique) and the percentage of test set attribute values included in the sampled set (Seen).
Table \ref{tab:prompt_engineering_with_example_values} shows that providing too few (3) or too many (10) example values harms GPT-3.5's performance, while GPT-4 maintains consistent F1-scores, peaking with 5 example values.
The sampled attribute values represent only 6\% to 28\% of all unique test set values, indicating that simply looking up these values is not enough to achieve F1-scores above 70\%. Both models derive general patterns from the \texttt{json-val} target schema representation.
Including example values increases the number of tokens and associated costs. Comparing costs per 1k extracted attribute-value pairs of GPT-3.5 and GPT-4 to the shortest \texttt{list} prompt template, \texttt{json-val} prompts are three times more expensive.

\begin{table}[]
\centering
\caption{F1-scores for prompt templates with example values. The highest F1-score per dataset is marked in bold.}
\label{tab:prompt_engineering_with_example_values}
\begin{tabular}{@{}l|rr|rr|rr|rr@{}}
\toprule
            & \multicolumn{4}{l|}{OA-Mine}                                                                                                           & \multicolumn{4}{l}{AE-110K}                                                                                                           \\ \midrule
Prompt      & Unique & Seen & GPT-3.5 & GPT-4 & Unique & Seen & GPT-3.5 & GPT-4 \\ \midrule
list        & 0                                    & 0\%                                  & 63.3                        & 69.1                      & 0                                    & 0\%                                  & 61.4                        & 56.34                     \\
json-3-val  & 283                                  & 6\%                                  & 74.4                        & 74.81                     & 172                                  & 20\%                                 & 66.6                        & 69.49                     \\
json-5-val  & 429                                  & 9\%                                  & 74.0                        & 74.55                     & 222                                  & 24\%                                 & 69.7                        & 72.76                     \\
json-10-val & 719                                  & 13\%                                 & 69.8                        & 75.02                     & 271                                  & 28\%                                 & 74.4                        & 70.08                     \\ \bottomrule
\end{tabular}
\end{table}

\subsection{In-Context Demonstrations}
\label{sec:in_context_learning} 
This section examines the impact of adding in-context learning demonstrations from the training set to the prompt templates from three perspectives: (i) the number of demonstrations selected, (ii) the effect of training set size on LLM performance, and (iii) a comparison of GPT and Llama-3 LLMs.

\begin{figure}[ht]
\centering
\includegraphics[width=.9\textwidth]{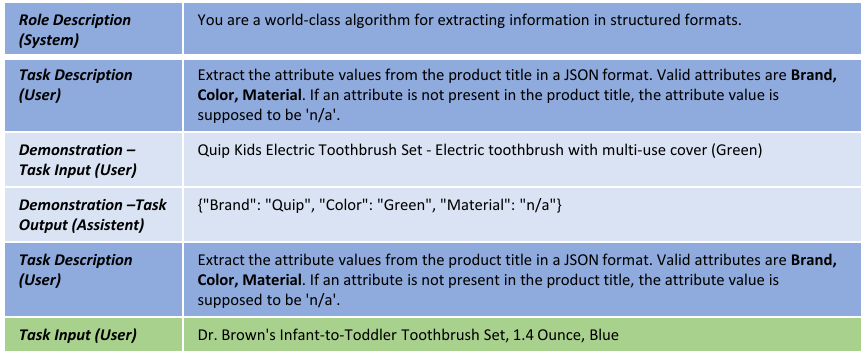}
\caption{Prompt template for in-context learning.}
\label{fig:few_shot}
\end{figure}

\vspace{.1cm}\noindent\textbf{Prompt Templates.}
The prompt templates \texttt{list} and the \texttt{schema}, as introduced in Section \ref{sec:prompt_engineering}, are extended to include demonstrations (\texttt{dem}).
Figure \ref{fig:few_shot} illustrates this extension.
Each demonstration (\textcolor{cyan}{light blue}) consists of a task input and a task output. The demonstrations are added to the chat prompt following the role and task description (\textcolor{blue}{blue}). The task description (\textcolor{blue}{blue}) is then repeated, followed by the task input (\textcolor{green}{green}). Task input and output of the demonstration are of the message types \texttt{user} and \texttt{assistent}, respectively.

\vspace{.1cm}\noindent\textbf{Amount of Demonstrations.}
The first dimension we explore is the number of demonstrations selected from the training set. 
We evaluate GPT-3.5 using the prompt template \texttt{list-dem} with 3, 10 and 15 demonstrations. To find semantic similarity demonstrations, the product titles of the training demonstrations are embedded using OpenAI's embedding model \texttt{text-embedding-ada-002}\footnote{https://platform.openai.com/docs/guides/embeddings/}. The embedded demonstrations with the greatest cosine similarity to the embedded task input are considered to be semantically similar.
Table \ref{tab:in_context_no_demonstrations} shows that the highest F1-score of 79.9\% is achieved with 10 demonstrations, which is 2\% better than with 3 demonstrations. Usage fees for hosted LLMs like GPT-3.5 depend on the number of tokens in the input prompts. Increasing the number of tokens by adding demonstrations can significantly raise usage fees. The right columns in Table \ref{tab:in_context_no_demonstrations} illustrate the cost for extracting one thousand attribute values based on December 2023 OpenAI usage fees\footnote{https://openai.com/pricing}. Using 10 demonstrations is twice as expensive as using 3, and using 15 demonstrations increases the cost further without improving the F1-score.

\begin{table}[ht]
\centering
\caption{F1-scores and extraction cost of the \texttt{list-dem} prompt template with different amounts of demonstrations. The highest F1-score per dataset is marked in bold.}
\label{tab:in_context_no_demonstrations}
\begin{tabular}{@{}r|rr|rr|rr@{}}
\toprule
 & \multicolumn{2}{l|}{OA-Mine}                            & \multicolumn{2}{l|}{AE-110K}                            & \multicolumn{2}{l}{Average}          \\ \midrule
 &    & \$ for 1k &    & \$ for 1k &         & \$ for 1k \\
 & F1 & A/V pairs & F1 & A/V pairs & F1      & A/V pairs \\ \midrule
%1                    &      74.8              & 0.0520                        &   76.2                 & 0.0507                        &   75.5                     & 0.0513                        \\
3                    &      77.0              & 0.0746                        &     79.1              & 0.0664                        &           78.1            & 0.0705                        \\
%5                    &      77.3            & 0.0978                        &   81.5                & 0.0835                        &     79.4                    & 0.0907                        \\
10                   &      \textbf{77.5}              & 0.1558                        &     \textbf{82.2}              & 0.1259                        &       \textbf{79.9}                  & 0.1408                        \\
15                   &      76.9              & 0.2133                        &    82.2                & 0.1643                        &    79.6                    & 0.1888                        \\ \bottomrule
\end{tabular}
\end{table}

\vspace{.1cm}\noindent\textbf{Amount of Training Data.}
We evaluate the training data efficiency of LLMs by assessing the \texttt{list-dem} prompt with 10 demonstrations from both small (S) and large (L) training sets for all LLMs. For GPT-4, we also report results for the \texttt{json-val-dem} prompt, as GPT-4 benefits from extended attribute representation and demonstrations.
Table \ref{tab:in_context_different_models} shows a marginal F1-score gain when demonstrations are sampled from the larger training set. GPT-4 gains 2\% with the large training set using the \texttt{json-val-dem} prompt, achieving the best average F1-score of 85\%.
In the following, we refer to the combination of the \texttt{json-val-dem} prompt and GPT-4 as ExtractGPT.

\begin{table}[t]
\centering
\caption{F1-scores for selecting demonstrations from either the small training set (S) or the large training set (L). The highest F1-score per dataset is marked in bold.}
\label{tab:in_context_different_models}
\begin{tabular}{@{}l|l|rr|rr|rr@{}}
\toprule
 &                    & \multicolumn{2}{l|}{OA-Mine}   & \multicolumn{2}{l|}{AE-110K}   & \multicolumn{2}{l}{Average}   \\ 
Prompt      & Model          & S       & L & S       & L & S       & L \\ \midrule
list-dem & GPT-3.5              & 77.5                        & 77.7                  & 82.2                       & 82.3                 & 79.9                        & 80.0                  \\
& GPT-4                & 78.2                        & 78.0                  & 80.4                        & 82.6                  & 79.3                        & 80.3                  \\
& Llama-3-8B               & 79.0                        & 77.9                 &  81.7                        & 81.8  &  80.4                       & 79.8                  \\  
& Llama-3-70B               &   81.8                      & 81.5                 &  84.0            & 84.1  &  82.9                       & 82.8                 \\ 
%& Beluga-7B           & 59.0                        & 59.0                  & 79.9                        & 80.0                  & 69.4                        & 69.5                  \\
%& Beluga2             &  65.7       & 65.8                  & 84.5                        & 84.5                  & 75.1                        & 75.2                  \\
%& SOLAR               & 63.9                        & 63.8                  &  83.5        &  83.3  & 63.9                        & 63.8                  \\   
\midrule
json-val-dem & GPT-4 & 80.2 & \textbf{82.2} & 85.5 & \textbf{87.5} & 82.8 & \textbf{84.9} \\
% & Llama-3-70B & 78.7 & 81.5 & 83.0 & 84.5 &  & 83.0 \\
  \bottomrule
\end{tabular}
\end{table}

\vspace{.1cm}\noindent\textbf{Comparison to Llama-3 LLMs.} 
Table \ref{tab:in_context_different_models} also reports results for Llama-3-8B and Llama-3-70B. 
The best-performing open-source LLM, Llama-3-70B, is only 2\% worse than GPT-4, making it a competitive alternative if sufficient GPUs are available. Llama-3-8B is a more compute-efficient option, with an average F1-score only 3\% lower than Llama-3-70B.
All LLMs achieve higher F1-scores on AE-110K than on OA-Mine, indicating that OA-Mine is more challenging than AE-110K. This is because OA-Mine has four times more unique attribute values than AE-110k and 82\% of the unique values in OA-Mine's test set are unseen, compared to 71\% in AE-110K.
%According to the results, GPT-4 performs better on OA-Mine, indicating that it learns more general extraction rules. On the other hand, the open-source LLMs are better at detecting similar values between the training and test sets, as demonstrated by their performance on AE-110K.

\subsection{Fine-Tuning}
\label{sec:fine_tuning} 
In this section, we evaluate the impact of fine-tuning on the AVE performance of GPT-3.5. First, we compare the fine-tuned GPT-3.5 models with Extract-GPT from Section \ref{sec:in_context_learning}. Second, we examine if fine-tuning imparts general knowledge useful for extracting attribute values from products not in the training set.

\vspace{.1cm}\noindent\textbf{Procedure.}
GPT-3.5 is fine-tuned using the training sets of OA-Mine and AE-110K. The training records are formatted with \texttt{list} and \texttt{json-val} templates, generating role descriptions, task descriptions, and task inputs, with task outputs containing attribute-value pairs. For \texttt{json-val}, 10 example values per attribute are sampled from the training set. These pre-processed datasets are uploaded to OpenAI's fine-tuning API\footnote{https://platform.openai.com/docs/guides/fine-tuning} and GPT-3.5 is fine-tuned for three epochs on these datasets using the OpenAI's default parameters.

%\begin{table}[h]
%\caption{Fine-tuning cost in \$ per prompt and dataset.}
%\label{tab:ft_cost}
%\begin{tabular}{@{}l|lr@{}}
%\toprule
%            & OA-Mine                  & AE-110K} \\ \midrule
%list        & \multicolumn{1}{r}{9.6}  & 7.2                         \\
%json-val & \multicolumn{1}{r}{33.6} & 28.8                        \\ \bottomrule
%\end{tabular}
%\end{table}

\vspace{.1cm}\noindent\textbf{Performance.}
We evaluate the F1-scores and extraction costs of fine-tuned models compared to ExtractGPT.
Table \ref{tab:finetuning_model_comparison} shows that fine-tuned GPT-3.5 models achieve similar F1-scores to ExtractGPT, averaging 85\%.
Fine-tuned GPT-3.5 models significantly reduce API usage costs. Extracting 1k attribute-value pairs costs up to 70 times more with ExtractGPT than with the fine-tuned GPT-3.5 models.
Considering fine-tuning and extraction costs, it is more cost-effective to fine-tune GPT-3.5 with the \texttt{list} prompt on the large training set, starting with about 500 attribute-value pairs, than to use ExtractGPT.

\begin{table}[]
\centering
\caption{F1-scores and the extraction cost (\$ for 1k A/V pairs) of  fine-tuned GPT-3.5 models and GPT-4. The highest F1-score per dataset is marked in bold.}
\label{tab:finetuning_model_comparison}
\begin{tabular}{@{}l|l|lr|lr@{}}
\toprule
              &             & \multicolumn{2}{l|}{OA-Mine}                              & \multicolumn{2}{l}{AE-110k}                              \\ \midrule
                &            &                          & \$ for 1k &                          & \$ for 1k \\
Prompt          & LLM      & F1                       & A/V pairs & F1                       & A/V pairs \\ \midrule
json-val-10-dem & GPT-4      & \multicolumn{1}{r}{82.2} & 6.4411                        & \multicolumn{1}{r}{87.5} & 6.3123                        \\
list            & ft-GPT-3.5 & \multicolumn{1}{r}{83.6} & 0.2600                        & \multicolumn{1}{r}{85.7} & 0.2643                        \\
json-val        & ft-GPT-3.5 & \multicolumn{1}{r}{84.5} & 1.2307                        & \multicolumn{1}{r}{86.0} & 0.9709                        \\ \bottomrule
\end{tabular}
\end{table}

\vspace{.1cm}\noindent\textbf{Generalization.}
The following analysis examines whether fine-tuned GPT-3.5 models can generalize to products and their attribute values not included in the training data. This is a crucial requirement due to the constant emergence of new products.
We test this by applying the two models fine-tuned on OA-Mine to AE-110k, and vice versa. These four fine-tuned models are compared to a plain GPT-3.5 as a baseline.
The results in Table \ref{tab:ft_transfer} indicate the fine-tuned models perform, on average, 17\% worse than the plain GPT-3.5, suggesting they may not generalize well and may lose some general language comprehension skills.
Based on these observations, we identify two scenarios for users employing GPT-3.5 or GPT-4 for AVE: (i) For frequent AVE on a known set of products, the lower token usage fee of the fine-tuned model justifies the fine-tuning cost. (ii) For infrequent AVE with constantly changing products, ExtractGPT is recommended.

\begin{table}[ht]
\centering
\caption{F1-scores of the fine-tuned GPT-3.5 models transferred to the other dataset.}
\label{tab:ft_transfer}

\begin{tabular}{@{}l|l|rr@{}}
\toprule
Dataset & Model                         & list & json-val \\ \midrule
OA-Mine & GPT-3.5                       & 63.3                     & 69.8                            \\
        & ft-GPT-3.5 on AE-110k & 43.7                     & 53.9                            \\ 
%        \cmidrule{2-4}
%        & $\Delta_1$ (ft-GPT-3.5 vs. GPT-3.5)                         & -19.6                    & -15.9                           \\ 
\midrule
AE-110K & GPT-3.5                       & 61.4                     & 74.4                            \\
        & ft-GPT-3.5 on OA-Mine & 46.2                     & 55.9                            \\ 
%        \cmidrule{2-4}
%        & $\Delta_2$ (ft-GPT-3.5 vs. GPT-3.5)                         & -15.2                    & -18.5                           \\ 
\bottomrule
\end{tabular}
\end{table}

\section{Comparison to PLM-based Baselines}
\label{sec:baselines}
This section compares the performance of ExtractGPT to the PLM-based baselines SU-OpenTag\footnote{https://github.com/hackerxiaobai/OpenTag\_2019/tree/master}~\cite{xu_scaling_2019}, AVEQA~\cite{wang_learning_2020}, and MAVEQA\footnote{https://github.com/google-research/google-research/tree/master/mave}~\cite{yang_mave_2022}. 
To assess training data efficiency, we fine-tune the baseline methods on both the small (S) and large (L) training sets. We ensure a fair comparison to ExtractGPT by using the same training sets for in-context learning.

\vspace{.1cm}\noindent\textbf{Absolute Performance.}
Table \ref{tab:baseline_comparison} shows the F1-scores of the fine-tuned PLM-based baselines and ExtractGPT.
The best average F1-score of SU-OpenTag and AVEQA is 6\% lower than the average performance of ExtractGPT.

\begin{table}[ht]
\centering
\caption{F1-scores of PLM baselines and GPT-4 on all and unseen attribute-value pairs in the test sets. The highest F1-score per dataset is marked in bold.}
\label{tab:baseline_comparison}
\begin{tabular}{@{}l|rr|rr|rr||rr|rr|rr@{}}
\toprule
            & \multicolumn{6}{l||}{All Attribute-Value Pairs}                                                                                    & \multicolumn{6}{l}{Unseen Attribute-Value Pairs}                                                                                                   \\ \midrule
            & \multicolumn{2}{l|}{OA-Mine}                   & \multicolumn{2}{l|}{AE-110K}                   & \multicolumn{2}{l||}{Average} & \multicolumn{2}{l|}{OA-Mine}                   & \multicolumn{2}{l|}{AE-110k}                   & \multicolumn{2}{l}{Average}                   \\ \midrule
            & S & L & S & L & S            & L            & S & L & S & L & S & L \\ \midrule
SU-OpenTag  & 55.1                  & 73.9                  & 70.6                  & 85.5                  & 62.8         & 79.7         & 42.0                  & 58.4                  & 32.0                  & 40.9                  & 37.0                  & 49.6                  \\
AVEQA       & 67.0                  & 78.7                  & 76.8                  & 80.9                  & 71.9         & 79.8         & 56.6                  & 65.0                  & 49.8                  & 49.3                  & 53.2                  & 57.2                  \\
MAVEQA      & 23.1                  & 65.7                  & 57.7                  & 76.8                  & 40.4         & 71.3         & 18.8                  & 42.6                  & 6.2                   & 24.9                  & 12.5                  & 33.7                  \\ \midrule
Extract-GPT & 80.2 & \textbf{82.2} & 85.5 & \textbf{87.5} & 82.8 & \textbf{84.9}  & \textbf{76.4}                  & 74.7                  & \textbf{68.8}                  & 59.1                  & \textbf{72.6}                  & 66.9                  \\ \bottomrule
\end{tabular}
\end{table}

\vspace{.1cm}\noindent\textbf{Training Data-Efficiency.}
Comparing PLM-based baselines fine-tuned on small and large training sets reveals a performance gap of 8\% to 31\%, indicating their need for large amounts of training data. In contrast, ExtractGPT shows only a 2\% performance gap between small and large training sets. Notably, ExtractGPT using the small training set outperforms all PLM-based methods using the large training set, despite having a training set five times smaller.

\vspace{.1cm}\noindent\textbf{Unseen Attribute Values.}
Our final analysis examines how PLM-based models and ExtractGPT perform with unseen attribute values. This is crucial for handling new products in e-commerce. We compare performance differences on test set attribute values not included in the training sets. OA-Mine has 1,607 and 1,073 unseen pairs, while AE-110K has 572 and 414, depending on the training set size.
Table \ref{tab:baseline_comparison} shows that all methods struggle with unseen values because lower F1-scores compared to the scenario with all attribute-value pairs are achieved. ExtractGPT is more robust, achieving an average F1-score 19\% higher than the best PLM-based method, AVEQA, on the small training set.

\section{Conclusions}

This paper explores the use of Large Language Models (LLMs) for product attribute value extraction (AVE). It evaluates various zero-shot and few-shot prompt templates with different LLMs. The best overall performance was achieved by GPT-4, with an F1-score of 85\%, using a prompt incorporating schema knowledge, attribute descriptions, and demonstrations. Notably, GPT-4's performance is only 3\% higher than the performance of Llama-3-70B, an open-source LLM, highlighting its competitiveness. Our experiments showed that LLMs are more training data efficient for AVE compared to PLMs. Given the same training data, GPT-4 surpassed the best PLM-based methods by 6\% in F1-score and proved more robust to unseen attribute values. Fine-tuning GPT-3.5 increased its performance close to GPT-4's, but reduced GPT-3.5's ability to generalize to new product attribute values not included in the training data.

\bibliographystyle{splncs04}
\bibliography{library}

\end{document}